\documentclass{article}

\usepackage[a4paper, total={6in, 8in}]{geometry}
\usepackage{wrapfig}
\usepackage{enumerate}
\usepackage{placeins}
\usepackage[title]{appendix}

\clearpage
\usepackage[utf8]{inputenc}

\usepackage{hyphenat}
\usepackage[main=english,german]{babel}
\usepackage{biblatex}

\usepackage{csquotes}  %
\usepackage[T1]{fontenc}
\usepackage{xspace}
\usepackage{textcomp}
\usepackage{mparhack}
\usepackage{relsize}
\usepackage{footnote}

\usepackage[linesnumbered,ruled,vlined]{algorithm2e}

\let\origcite\cite%
\def\cite#1{\unskip~\origcite{#1}}

\usepackage{amsmath}
\usepackage{bbold}
\usepackage{amssymb}
\DeclareSymbolFontAlphabet{\amsmathbb}{AMSb}%
\usepackage{upgreek}
\usepackage{bbm}
\usepackage{bm}      %
\usepackage{mathtools}
\usepackage{isomath}
\usepackage{dutchcal}
\usepackage{amsthm}

\newcommand{\approptoinn}[2]{\mathrel{\vcenter{
  \offinterlineskip\halign{\hfil$##$\cr
    #1\propto\cr\noalign{\kern2pt}#1\sim\cr\noalign{\kern-2pt}}}}}

\usepackage{wrapfig}

\usepackage{array}
\usepackage{tabularx}
\usepackage{makecell}
\usepackage{multirow}
\usepackage{booktabs}
\newcolumntype{C}[1]{>{\centering\arraybackslash}p{#1}}
\newcolumntype{R}[1]{>{\raggedleft\arraybackslash}p{#1}}
\newcolumntype{L}[1]{>{\raggedright\arraybackslash}p{#1}}
\newcolumntype{a}{>{\hsize=.70\hsize}X}
\newcolumntype{B}{>{\hsize=.90\hsize}X}
\newcolumntype{d}{>{\hsize=.20\hsize}X}
\newcolumntype{e}{>{\hsize=.40\hsize}X}
\newcolumntype{z}{>{\hsize=.10\hsize}X}
\newcolumntype{q}{>{\raggedright\arraybackslash}X}
\newcolumntype{f}{>{\hsize=.50\hsize}X}
\newcolumntype{N}{>{\centering\arraybackslash}X}
\newcolumntype{Y}{>{\centering\arraybackslash}X}
\newcolumntype{y}{>{\raggedleft\arraybackslash}X}
\newcolumntype{g}{>{\centering\arraybackslash}X}
\newcolumntype{G}{>{\hsize=.5\hsize}r}

\usepackage{ltablex}
\usepackage{subfig}
\DeclareCaptionLabelFormat{lc}{\MakeLowercase{#1}~#2}
\usepackage{blindtext}

\usepackage[dvipsnames]{xcolor}
\definecolor{dpurple}{HTML}{5e3c99}
\definecolor{lpurple}{HTML}{b2abd2}
\definecolor{dorange}{HTML}{e66101}
\definecolor{lorange}{HTML}{fdb863}
\definecolor{llpurple}{HTML}{DFD8FF}

\usepackage[%
  hyperfootnotes=false,%
  pdfpagelabels,%
]{hyperref}
\usepackage{hyperxmp}
\usepackage{graphicx}
\usepackage[export]{adjustbox}
\usepackage{rotating}

\usepackage{tikz}

\usetikzlibrary{arrows,shapes,automata,backgrounds, arrows.meta}
\usetikzlibrary{positioning}
\usetikzlibrary{decorations.pathreplacing}

\tikzset{entity/.style={
    draw=#1,
    thick,
    ellipse,
    minimum width=0.75cm,
    minimum height=0.75cm,
    font=\small,
    outer sep=3pt,
  },
  text style/.style={
    sloped,
    text=black,
    font=\footnotesize,
    above
  }
}

\tikzset{entity node/.style={entity,
                            draw=lorange!75,
                            fill=lorange!20,
                            align=center,
                            font=\bfseries,
                            dotted,
                            line width = 1.2mm
                            }
}
 \tikzset{literal node/.style={entity,
                                draw=dorange,
                                fill=dorange!60,
                                align=center,
                                inner sep=2pt,
                                font=\bfseries,
                                text=white
                                }
}

\tikzset{scalar node/.style={entity,
                             draw=lpurple!75,
                             fill=lpurple!60,
                             align=center,
                             font=\bfseries,
                             }
}

\tikzset{calculated node/.style={entity,
                                 thick,
                                 draw=dpurple,
                                 fill=dpurple!60,
                                 align=center,
                                 font=\bfseries,
                                 text=white
                            }
}

\tikzset{edgestyle/.style={ fill=white,
                            anchor=center,
                            pos=0.5,
                            font=\bfseries
                            }
}
\tikzset{isa/.style={       -{Latex[open]},
                            dashed,
                            shorten >=1pt,
                            auto
                            }
}
\tikzset{isalabel/.style={
                            fill=white,
                            anchor=center,
                            pos=0.4
                            }
}

\usepackage{pgfplots}
\pgfplotsset{compat=newest}
\usetikzlibrary{shapes.geometric}

\usepackage[capitalise]{cleveref}

\crefname{algocf}{Algorithm}{Algorithms}
\Crefname{algocf}{Algorithm}{Algorithms}
\Crefname{prop}{Proposition}{Propositions}
\Crefname{theorem}{Theorem}{Theorems}
\Crefname{corollary}{Corollary}{Corollaries}
\Crefname{lemma}{Lemma}{Lemmas}
\Crefname{subsubsection}{Section}{Sections}
\crefname{subsubsection}{section}{sections}
\crefname{chapter}{Chapter}{Chapters}
\Crefname{chapter}{Chapter}{Chapters}
\crefname{part}{Part}{Parts}
\crefname{part}{Part}{Parts}
\Crefname{equation}{Relation}{Relations}
\crefname{equation}{Relation}{Relations}
\Crefname{figure}{Figure}{Figures}
\Crefname{figure}{Figure}{Figures}
\Crefname{table}{Table}{Tables}

\PassOptionsToPackage{printonlyused}{acronym}
\usepackage{acronym}

\usepackage{enumitem}

\usepackage{siunitx}

\newcommand\vmath[1]{\ensuremath{#1}\xspace}

\newcommand\commentOut[1]{}

\newcommand\vcomment[1]{}

\def\@fnsymbol#1{\ensuremath{\ifcase#1\or *\or \dagger\or \ddagger\or
  \mathsection\or \mathparagraph\or \|\or **\or \dagger\dagger
  \or \ddagger\ddagger \else\@ctrerr\fi}}
\newcommand{\ssymbol}[1]{^{\@fnsymbol{#1}}}

\DeclareMathOperator*{\argmax}{arg\,max}

\newcommand\tv[1]{{\pmb{\MakeLowercase{#1}}}}
\newcommand{\norm}[1]{\left\lVert#1\right\rVert}

\Crefname{table}{Table}{Tables}

\SetKwProg{Fn}{Function}{}{}
\SetKwInOut{Parameter}{Parameters}
\SetKw{KwVar}{Var}
\SetKw{KwSet}{Set}
\SetKw{KwList}{List}
\SetKw{KwInit}{Init:}
\SetKw{KwParams}{Params:}
\SetKw{KwTuple}{Tuple}
\SetKw{KwBreak}{break}
\SetKw{KwNot}{Not}

\SetCommentSty{mycommfont}

\SetAlFnt{\sffamily}

\makeatletter
\renewcommand{\algocf@Vline}[1]{%
  \strut\par\nointerlineskip%
  \algocf@push{\skiprule}%
  \hbox{\bgroup\color{ChapBlue}\vrule\egroup%
    \vtop{\algocf@push{\skiptext}%
      \vtop{\algocf@addskiptotal #1}\bgroup\color{ChapBlue}\Hlne\egroup}}\vskip\skiphlne%
  \algocf@pop{\skiprule}%
  \nointerlineskip}%
\renewcommand{\algocf@Vsline}[1]{%
  \strut\par\nointerlineskip%
  \algocf@bblockcode%
  \algocf@push{\skiprule}%
  \hbox{\bgroup\color{ChapBlue}\vrule\egroup%
    \vtop{\algocf@push{\skiptext}%
      \vtop{\algocf@addskiptotal #1}}}%
  \algocf@pop{\skiprule}%
  \algocf@eblockcode%
}
\makeatother

\newcommand\ts[1]{\vmath{\mathrm{\MakeUppercase{#1}}}}

\newcommand{\matr}[1]{\MakeUppercase{\mathbf\mathit\bm{#1}}\protect}
\clearpage

\bibliography{refs}%

 \author{%
 Thomas Asikis
 }

\usepackage[utf8]{inputenc}

\title{Multi-Objective Optimization for Value-Sensitive and Sustainable Basket Recommendations}

\makeatletter
\renewcommand\@date{{%
  \vspace{-\baselineskip}%
  \large\centering
  \begin{tabular}{@{}c@{}}
    \normalsize asikist@ethz.ch \\
    \normalsize asikis@soziologie.uzh.ch
  \end{tabular}%

  \bigskip

  \textsuperscript{1}Chair of Computational Social Science, ETH Zurich, Zurich, Switzerland
  \\
  \textsuperscript{2}Game Theory, UZH Zurich, Zurich, Switzerland

  \bigskip
}}
\makeatother
\begin{document}
\maketitle
\begin{abstract}
   Sustainable consumption aims to minimize the environmental and societal impact of the use of services and products. Over-consumption of services and products leads to potential natural resource exhaustion and societal inequalities as access to goods and services becomes more challenging. In everyday life, a person can simply achieve more sustainable purchases by drastically changing their lifestyle choices and potentially going against their personal values or wishes. Conversely, achieving sustainable consumption while accounting for personal values is a more complex task as potential trade-offs arise when trying to satisfy environmental and personal goals. This article focuses on value-sensitive design of recommender systems, which enable consumers to improve the sustainability of their purchases while respecting personal and societal values. Value-sensitive recommendations for sustainable consumption are formalized as a multi-objective optimization problem, where each objective represents different sustainability goals and personal values. Novel and existing multi-objective algorithms calculate solutions to this problem. The solutions are proposed as personalized sustainable basket recommendations to consumers. These recommendations are evaluated on a synthetic dataset, which comprises three established real-world datasets from relevant scientific and organizational reports. The synthetic dataset contains quantitative data on product prices, nutritional values, and environmental impact metrics, such as greenhouse gas emissions and water footprint. The recommended baskets are highly similar to consumer purchased baskets and aligned with both sustainability goals and personal values relevant to health, expenditure, and taste. Even when consumers would accept only a fraction of recommendations, a considerable reduction of environmental impact is observed.
\end{abstract}

\section{Introduction}
Sustainable Development Goals~\cite{Uni15} (SDG) have been proposed by the United Nations (UN) to describe several sustainability criteria in form of goals, tasks, and scenarios.
Environment and society related sustainability goals can be broken down to numerous constraints and objectives~\cite{manzoor2014optimal,Chaudhary2018} that affect everyday decisions, 
such as the reduction of CO\textsubscript{2} emission with everyday activities.
Making a desired decision with respect to such constraints and objectives often introduces potential trade-offs, e.g. making a decision that minimizes CO\textsubscript{2} emissions of a certain activity may increase the corresponding water footprint.
Including personal values, such as preference/taste or budget constraints may introduce even more trade-offs and make the decision process more challenging.
Several decision-support methods that tackle such problems are considered under the umbrella term of Artificial Intelligence (AI)~\cite{glasmachers2010natural,Marler2004,abejon2020multi}.
AI is expected to play a key role~\cite{vinuesa2020role} in solving complex challenges arising from SDGs.
Nevertheless, applying AI may also inhibit sustainability challenges relevant to value\hyp{}sensitive design, namely challenges related to: 
(i) unequal distribution of resources, (ii) loss individual of autonomy and privacy, and (iii) increased emissions from calculations~\cite{vinuesa2020role,Helbing2015}.
Value\hyp{}sensitive design of AI systems aims to address such challenges~\cite{van2007ict}, by including personal and societal values as objective or constraints of the AI system.

Existing macroscopic simulation and control models are used to estimate environmental impact but often do not deal or represent real-world data on individual/microscopic level.
Such models mostly rely on theoretical economical and climate models, which evaluate sustainability on a macroscopic or societal level~\cite{manzoor2014optimal,traeger20144}.
A welfare objective function that includes analytical terms representing satisfaction from consumption and environmental pollution is often optimized~\cite{nordhaus1993optimal,traeger20144}.
The optimization outcomes of such models are then used in analysis and policy making.
Although macroscopic and/or centralized models can provide useful insights on decision-making at organizational level, 
it may not be possible to anticipate for errors in estimation of individual objective functions and system state.
Individual consumption is one of the main driving factors of industrial production, and thus has great environmental and societal impact~\cite{figge2014sufficiency}.
Every day, consumers are urged to take several decisions to fulfill their personal goals and values, mainly originating from individual needs and wishes.
Criticism on centralized models often focuses on poor practical applications and potential high estimation errors~\cite{botzen2012sensitive}, as well as potential challenges to addressing personal values and morals~\cite{dignum2019responsible}.
It may prove rather challenging to simulate and estimate individual or personalized recommendations that assist sustainable decision-making in a centralized macroscopic manner.
Data-oriented approaches that focus on individual/microscopic level can be directly applied in real-world scenarios in the form of mobile applications and website recommender systems~\cite{asikis2020value}, 
which can assist individuals in achieving more sustainable consumption~\cite{su132011162} consciously.
Relevant recommender systems and analyses mainly focus on diets or single product recommendations~\cite{abejon2020multi,tomkins2018sustainability,starke2019recsys}.

Shopping decisions often aim to satisfy several criteria, such as the taste, the ability of purchased products to be combined together, available consumer budget, and health implications of consuming the product.
Trade-offs arise when optimizing for personal values, e.g. when a consumer needs to decide between a healthier and cheaper product.
Such trade-offs are mainly addressed by multi-objective optimization in modern recommender systems~\cite{zuo2015personalized,ZHENG2022141,pachot:hal-03349092,patro2020towards}.
Value\hyp{}sensitive design can be applied to create systems that support individual decisions when resolving these trade-offs~\cite{asikis2020value}.
Following a value\hyp{}sensitive design indicates that personal values can be incorporated in the system design, often introducing new constraints and objectives to the optimization.
Thus, combining value\hyp{}sensitive design and sustainable decision\hyp{}making results in even more complex and challenging trade-off optimization problems.
Although the effect of sustainable recommendations on individual purchases is often observed~\cite{nghiem2016mobile,asikis2020value}, there is seemingly less quantitative analysis on the collective impact of more sustainable decisions on emissions, pollution, resource usage, and personal values.

The main contributions of the current article are to: (i) propose and formalize a real-world multi-objective optimization problem for recommendations of personalized sustainable baskets, (ii) create and analyze a novel synthetic dataset based on sustainability and consumer real\hyp{}world data, (iii) propose effective existing baselines for this problem, and (iv) design and implement a novel deep learning framework for mixed integer programming and multi-objective optimization problems.
In \cref{section:methods:problem} an explicit formalization of the proposed optimization problem is presented. 
Therein constraints, trade-offs, and objectives originating from value\hyp{}sensitive design and SDGs are combined to form a multi-objective optimization problem.
The proposed dataset structure can be used to evaluate similar real-world datasets by online retailers to enable value\hyp{}sensitive sustainable recommendations.
\cref{section:methods:g3} introduces a novel deep learning architecture, termed gradient guided genetic algorithm (G3A), which combines the ability of evolutionary algorithms to solve complex multi-objective problems and the ability of Neural Ordinary Differential Equation Control (NODEC)\cite{asikis2020nnc,bottcher2022ai} to efficiently and timely control complex processes, such as genetic evolution.
\cref{sec:g3a:experimental:evaluation} presents the results of experimental evaluation of G3A and other multi-objective optimization algorithms on the problem discussed in  \cref{section:methods:problem}.
\cref{sec:g3a:conclusion} concludes the article.

\section{Preliminaries}\label{section:methods:problem}

\subsection{Problem Formulation}
Value\hyp{}sensitive sustainable recommendations are formalized as a multi-objective optimization problem of selecting combinations of discrete quantities over $N=132$ distinct products.
First, an intended basket is defined as the purchased weekly basket, i.e. a vector of non-negative integer product quantities $\boldsymbol{x}^*_{k,q} \in \mathbb{N}_0^N$ for a specific household $k$ at week $q$.
In a real-world application, where the purchased basket is not known, the consumer may provide an intended basket via a shopping list interface or an e-shop basket interface.
For brevity, week $q$ and household $k$ indices are omitted from the basket vector, as the proposed multi-objective optimization calculations do not require values of intended baskets that were purchased in the past or from other households.
The intended basket $\boldsymbol{x}^*$ is considered as the initial solution of the presented problem and is also considered as the basket that represents consumer taste and nutritional goals. 
For the current study an ordered set $C$ of $|C| = 11$ of possible recommendation features is considered, where a feature index $j=1$ indicates the corresponding feature as shown in~\cref{tab:g3a:features}.

\begin{table}[!ht]
    \centering
    \footnotesize
    \begin{tabularx}{\linewidth}{R{0.05\linewidth}L{0.3\linewidth}L{0.2\linewidth}L{0.15\linewidth}L{0.1\linewidth}}
    \toprule
    $j$ & Feature                   & Unit                 & Scope & Target \\
    \midrule
     1  & Cosine similarity         & -                                 & Personal    & Max. \\
     2  & Cost                      & Dollars (\$)                      & Personal    & Min. \\
     3  & Energy                    & kilo Calories (kCal)              & Personal    & Pres. \\
     4  & Protein                   & grams (g)                         & Personal    & Pres. \\
     5  & Fat                       & grams (g)                         & Personal    & Pres. \\
     6  & GHG emissions             & CO\textsubscript{2} kg eq. & Environment & Min. \\
     7  & Acidification  pollution  & SO\textsubscript{2} kg eq. & Environment & Min. \\
     8  & Eutrophication pollution  & PO\textsubscript{4}\textsuperscript{-3} kg eq. & Environment & Min. \\
     9  & Land use                  & m\textsuperscript{2}              & Environment & Min. \\
     10 & Water usage               & L                                 & Environment & Min. \\
     11 & Stressed water usage      & L                                 & Environment & Min.\\
    \bottomrule
    \end{tabularx}
    \caption[Features relevant to objectives for basket selection.]{
     Features relevant to objectives for basket selection. The first column shows the index of each feature, which also coincides with their position in the ordered set $C$. Feature, Unit, and Scope columns give a brief overview of the objectives and finally the Target column describes whether the goal of the optimization is to minimize, maximize, or preserve the intended basket value.
    }
    \label{tab:g3a:features}
\end{table}

The synthetic dataset provides coefficients $c_{i,j}$, which in this study are calculated based on the mean\footnote{Other estimators were also tested, such as the median values. Any measure of central tendency or the actual values at a current time and retailer store can be considered.} values over all transactions in the dataset and describe the corresponding feature $j$ quantity $c_{i,j}$ per unit for each product $i$.
Therefore, for a basket $\tv{x}$, one can calculate the total quantity for a specific feature as
\begin{equation}
\label{eq:g3a:basket:total}
    v_j(\boldsymbol{x}) = \sum_{i=1}^N c_{i,j}x_i.
\end{equation}
When designing the objective functions and comparing recommendations among baselines, often the ratio of  total feature quantities between two baskets $\tv{x},\tv{x}'$ 
\begin{equation}
\label{eq:g3a:basket:ratio}
    \rho_j(\tv{x}, \tv{x}') = \dfrac{v_j(\tv{x})}{v_j(\tv{x}')}
\end{equation}
is used.
In particular, most calculations related to environmental and personal objectives use the ratio of a recommendation towards the intended basket $\rho_j(\tv{x}, \tv{x}^*)$ for a specific feature $j$.

\subsubsection{Individual Objectives}

Consumer taste is the first personal value that is considered as an optimization objective, which is minimized when the recommended basket $\boldsymbol{x}$ is as similar as possible to the intended basket $\boldsymbol{x}^*$.
High similarity between a recommended basket $\tv{x}$ and the target intended $\boldsymbol{x}^*$ indicates higher likelihood of a purchase under a counterfactual hypothesis, in which the consumer would consider recommended baskets before purchase.
The first objective function to minimize depends on the cosine similarity between recommended and intended basket
\begin{equation}
\label{eq:sim:obj}
J_{1}\left(\boldsymbol{x},\boldsymbol{x}^*\right) = 1 - \dfrac{\boldsymbol{x}^{\top}\boldsymbol{x}^*}{\norm{\boldsymbol{x}}\norm{\boldsymbol{x}^*}}.
\end{equation}
\cref{eq:sim:obj} is minimized by recommending the intended basket.

The next personal value considered in optimization is a function of cost.
In general, it is assumed that individuals would prefer to minimize expenses and select cheaper baskets that satisfy their taste.
Next, the cost ratio between recommended and intended basket costs is calculated as an objective function:
\begin{equation}
    \label{eq:cost:obj}
    J_{2}\left(\boldsymbol{x},  \boldsymbol{x}^* \right) =   \rho_2(\boldsymbol{x},  \boldsymbol{x}^*).
\end{equation}
The intended basket does not minimize \cref{eq:cost:obj}, whereas a basket with no products at all would be an optimal solution.

Next, the nutritional values of a recommended basket are considered for optimization.
For each unit of product $i$ and nutritional product feature $j$ the nutritional quantity per unit $c_{i,j}$ is calculated.
Three nutritional features are denoted by indices $j\in\{3,4,5\}$.
The health objective functions use the intended basket nutritional value as a baseline to evaluate the difference for each nutritional feature between recommended and intended baskets:
\begin{equation}
    \label{eq:health:obj}
    J_{j}\left(\boldsymbol{x}, \hat{\boldsymbol{x}} \right) = \left(1-\rho_j(\tv{x},\tv{x}^*)\right)^2 =  \left(\dfrac{v_j(\boldsymbol{x}^*)-v_j(\boldsymbol{x})}{v_j(\boldsymbol{x}^*)}\right)^2, j \in \{3,4,5\}.
\end{equation}
The intended basket is one solution that minimizes the \cref{eq:health:obj}.

\subsubsection{Environmental Impact Objectives}

Collective environmental values are also considered based on the provided data from Ref.~\cite{poore2018reducing}.
In total, a set of six environmental impact criteria are considered for each product, as shown in~\cref{tab:g3a:features}, namely
\emph{green house gas} (GHG) emissions, which contribute to climate change,
\emph{acidifying} pollution that decreases fertility and can cause desertification, 
\emph{eutrophication} pollution, which destabilizes food chains in ecosystems, 
\emph{water usage} that has several environmental effects,
\emph{stress-weighted water} usage that takes into account whether the water is taken from arid/dry lands,
and \emph{land usage}, which is important to resource allocation for farming and deforestation.
The mean product features per unit are used as coefficients $c_{i,j}$ for calculating $v_{\text{j}}, j > 5$.
Similar to the price objective, the ratio between intended and recommended basket of each environmental impact feature is considered as an objective function:
\begin{equation}\label{eq:env}
J_{j}\left(\boldsymbol{x},\boldsymbol{x}^* \right) =\rho_j(\tv{x},\tv{x}^*)
\end{equation}
It is important to note that for the current dataset, there are no negative values for any coefficient $c_{i,j}$, thus all nominators and denominators of the proposed objectives are positive.
Unless the intended basket optimizes all of the above objectives simultaneously and is non-empty, then there is no solution that optimizes the above objectives simultaneously.
For example this can be shown when removing a single item from an non-empty intended basket.
The item removal will decrease the price objective value and also environmental impact objectives, while it will increase the taste objective value.

\subsubsection{Problem Formulation}

The proposed optimization framework evaluates the recommended baskets across a set $\ts{C} = \{1,2,...M\}$ of all \emph{M=11} different objectives presented above.
The optimization is performed in a decentralized manner and only uses the intended basket to decide objective function values.
The multi-objective problem for $M$ objectives can be summarized as
\begin{equation}
    \min_\tv{x} \left(J_1(\tv{x}, \tv{x}^*), J_2(\tv{x}, \tv{x}^*), ... J_M(\tv{x}, \tv{x}^*)\right), \quad
    \tv{x} \in \hat{\ts{X}}
\end{equation}
for a set of feasible baskets $\hat{\ts{X}}$.
An optimization algorithm $f(\ts{x}_0 ; \tv{w}) = \ts{x}$ with parameter vector $\tv{w}$ takes an initial set of baskets 
$\ts{x}_0$ and calculates a recommended set of baskets $\ts{x}$.
The goal of such algorithm is to find a non\hyp{}dominated set of baskets. 
A basket $\tv{x}$ dominates $\tv{x} \prec \tv{x}'$ another basket $\tv{x}'$ if $J_j(\tv{x}) \leq J_j(\tv{x}')$ for all $j\in C$ and $J_j(\tv{x}) < J_j(\tv{x}')$ holds at least for one $j$ ~\cite{deb2002fast}.
If no other basket dominates $\tv{x}$, then it is referred as non\hyp{}dominated.

The problem of selecting the optimal number of products that fill a consumer basket under budget and value-sensitive constraints may highly resemble a multi-objective and multi-dimensional knapsack problem~\cite{lust2012multiobjective}.
Solving such problem with constraint optimization and linear/dynamic programming may prove challenging, especially as it may not admit efficient Polynomial time approximations~\cite{kulik2010there} and linear relaxation methods may not yield desired solutions~\cite{lust2012multiobjective}.
Another widely applied approach using regularization techniques can also be considered to solve such problems and would require the fitting and interpretation of Lagrange coefficients of each objective~\cite{zare2020determination}, which resemble scalarization methods~\cite{ZHENG2022141}.
The article focuses mostly on the usage of evolutionary algorithms, such as Multi-Objective Natural Evolution Strategies (MO-NES)~\cite{glasmachers2010natural} and Reference point  Non\hyp{}dominated Sorting Genetic Algorithm II~\cite{deb2006reference} (RNSGA-II), which are shown to work on multi-objective optimization problems that have multiple ($>5$) objectives and the newly introduced G3A.

\subsection{Dataset}\label{supp:dataset}
Transaction data, product prices, and purchased quantities were retrieved by ``The Complete Journey Dataset'' by the Dunnhumby grocery store~\cite{DUNNHUMBY}.
The quantities are included in US imperial units and a conversion to metric system was done in the following manner: 
(i) Unit labels are identified and grouped together with regular expressions, e.g. "LB,lb, LBs" all represent the same label which denotes pounds. 
(ii) Weight and volumetric labels are separated and proper conversion coefficients are used to convert each unit to the corresponding metric unit used in the other datasets. 
(iii) Prices may change through time, so the mean price per unit is calculated through time and over all stores to generate the price features used by all algorithms.

Environmental impact indicators for product types are taken from Ref.~\cite{poore2018reducing}. 
Nutrition information from Food Agricultural Organization Food Balance Sheets~\cite{kelly1991food} are downloaded from Ref~\cite{FAOFBSO}.
All three datasets contain different product type labels for each product. 
From "The Complete Journey Dataset" the "SUB\_COMMODITY\_DESC" column is treated as the product identifier.
Each value of the column "SUB\_COMMODITY\_DESC" is matched against the "product category" column from FAO FBS dataset and the dataset column "product category" from Ref~\cite{FAOFBSO}.
The resulting dataset contains transaction prices, purchased quantities, environmental impact values, and nutritional info per transaction.

After processing, the dataset contains almost 885'000 transaction corresponding to  approximately 167'000  baskets purchased by almost 2'500 households.
The transactions were performed over a period of 709 days across 489 distinct stores. 
The FAO product categorization is used to identify products, and thus the product set is considered to contain 132 distinct products.
The proposed dataset\footnote{The relevant code will be made available publicly, but the release of the final dataset is pending confirmation from Dunnhumby.} is used to motivate and estimate environmental impact, consumer preferences, purchasing costs, and nutritional values of recommended baskets.
The proposed multi-objective recommender systems in the current article can also be applied to any dataset and basket selection problem that involves multiple-objectives relevant to product characteristics.

\section{Evolutionary Algorithms and Multi-Objective Optimization}

Multi-objective recommender systems are common in literature~\cite{ZHENG2022141,GENG2015383} and have been applied in sustainability related problems from producer perspective~\cite{pachot:hal-03349092} and recommendation of local businesses\cite{patro2020towards}.
Multi-objective evolutionary algorithms have been used to solve recommendation problems~\cite{GENG2015383}.
Evolutionary algorithms are widely used for multi-objective optimization~\cite{deb2000fast} with $M \geq 2$ objectives.
Although the current problem resembles a multi-task recommendation problem~\cite{ZHENG2022141} that is often treated with scalarization methods~\cite{ZHENG2022141}, the high number of available products to purchase, and the high number of personal and environmental values to consider makes evolutionary algorithms more suitable to implement.
Furthermore, this article further showcases how scalarization methods can be combined to evolutionary algorithms~\cite{glasmachers2010natural} to guide the evolution towards optimization of specific objectives.
A brief overview of evolutionary algorithms is illustrated in \cref{fig:g3:ga}.
Typically each basket, or solution\footnote{The term solution will be used in the sections that describe models in accordance to literature.} in the optimization context, $\tv{x}$ is mapped to an objective vector $\tv{\zeta}(\tv{x}) \in \mathbb{R}^M$, where each vector element represents an objective function value $\zeta_j = J_j(\tv{x})$.
Often, such algorithms improve a set of an initial population of solutions $\ts{x}_\tau$ by applying probabilistic operators on each solution vector $\tv{x}$, such as the \emph{random crossover}.
Random crossover randomly combines elements from different solutions $\tv{x},\tv{x}'$ with probability $p$
\begin{equation}
    x_i = \left\{
		\begin{array}{lr}
			x'_i & \mbox{if } \delta < p \\
			x_i & \mbox{otherwise } 
		\end{array}
	\right.
	,
\end{equation}
where $\delta$ is sampled from a probability density function $\delta\sim\mathcal{f}$ with finite support $[0,1]$.
Another probabilistic mechanism is the \emph{random mutation}, e.g. replacing an element of the solution with a random number sampled from a probability distribution $\kappa\sim\mathcal{f}_{\text{discrete}}$ to an element of the solution
\begin{equation}
    x_i = \kappa .
\end{equation}
Each new solution is evaluated based on the corresponding objective vector $\tv{\zeta}(\tv{x})$ and a \emph{selection} of solutions is performed.
Typically a non\hyp{}dominated sorting is performed for the selection of non\hyp{}dominated solution candidates both from new solutions and the initial population.
The non\hyp{}dominated sorting is performed recursively, i.e. each time a non\hyp{}dominated set $F_\alpha$ is selected, the non\hyp{}dominated solutions are assigned to $F_\alpha$ and then removed from the population. A new non\hyp{}dominated search is performed on the remaining solutions to determine the non\hyp{}dominated front $F_{\alpha+1}$.
This process repeats until all solutions are assigned to a front.
A possible selection mechanism would select all non\hyp{}dominated solutions, i.e. the solutions in $F_1$.
The selected solution candidates are preserved in a new population of solutions  $\ts{x}_{\tau + 1}$ and the whole process (crossover, mutation, selection) is repeated until a convergence criterion is met, e.g. no new solutions are preserved in a population after an iteration.
Often $\tau$ is referred to as a \emph{generation}.
A widely used algorithm that follows the above strategy for multi-objective optimization is the Non\hyp{}dominated Sorting Genetic Algorithm II (NSGA-II)~\cite{deb2002fast}.

\subsubsection{RNSGA-II}

A non\hyp{}dominated sorting algorithm may produce a large number of non\hyp{}dominated solutions that are not preferable, e.g. solutions that optimize a single objective very well and not the others.
To keep the population size $B$ per generation constant, a secondary selection operation needs to be performed.
Random selection is often undesired in problems that have multiple objectives~\cite{deb2006reference}, and thus a more sophisticated technique is preferred.
Some probabilistic evolutionary algorithms use a sorting operation to perform a secondary selection operation that guide the evolutionary processes towards preferred non\hyp{}dominated solutions, e.g. non\hyp{}dominated solutions that optimize specific combinations of the objectives very well.
A typical example that will be used as a baseline in the current study is reference point NSGA-II, abbreviated as RNSGA-II~\cite{deb2006reference}, which uses reference directions to guide evolution towards preferred solutions.
In brief, one or more reference points are selected to guide the evolution.
A reference point $\hat{\tv{\zeta}}$ is generated by providing a vector of preferred objective values to the system.
Each candidate solution receives two ranks determined by the non\hyp{}dominated sorting and a distance metric from each reference point, i.e. lower distance values receive lower ranks.
Lower ranks are used to select the candidates for the next generation.
This algorithm shows higher performance gains compared to NSGA-II to perform better on multi-objective problems with more than $2$ objectives~\cite{deb2006reference}.
RSNGA-II is used as a baseline in the current study following the default implementation of \cite{pymoo}.

A logistic map~\cite{fuertes2019chaotic} is applied on the initial basket to generate the initial solution for RNSGA-II, improving performance considerably compared to other random initialization.
Several reference points settings are tested for RNSGA-II.
The current reference points provided to RNSGA-II are three, one that is calculated by using the infeasible optimum, where every loss is 0, one that minimizes all individual losses (e.g. all values for $j\leq5$ are 0 and the rest are $1$), and one that minimizes all environmental losses (e.g. all values for $j>5$ are set to $0$).
Using less than 2 reference points resulted often in bad performance.
Other reference point settings were tested on $100$ intended baskets, such as using the one the intended basket or minimization of specific losses on smaller samples, but it was unclear whether better performance could be achieved by using them.
The current reference point setup was chosen, as it provided the best performing dominance ratio when comparing to other baselines.
Integer exponential crossover and polynomial mutation are used for the genetic operators.
Finally, other settings were tested with $B=100$, but were omitted the due to lower dominance ratio, slower convergence times, large number of solutions, and difficulty to determine subsets of good solutions.

\subsubsection{MO-NES}\label{chapter:appendix:g3:MONES}

Another way to handle multi-objective optimization problems is the use of MO-NES~\cite{glasmachers2010natural}, which use a gradient guided search algorithm to find non\hyp{}dominated solutions by parametrizing a probabilistic model (relies on sampling).
The algorithm optimizes the parameters of a model that samples solutions from underlying distributions.
For each solution, a sample vector $\tv{z} \in \mathbb{R}^N$ is generated, where each element is sampled from a normal distribution $z_i\sim \mathcal{N}(0,1)$.
A new solution $\tv{x}'$ is calculated based on a parent solution $\tv{x}' = \tv{x} +\sigma \matr{A}\tv{z}$, where $\sigma \in \mathbb{R}, \matr{A}\in \mathbb{R}^{N \times N}$ are the co-variance related terms.
Samples from the previous population $X_{\tau}$ and the new candidates $\tv{x}'$ are combined into an intermediate population $\ts{X}'$.
Each solution $\tv{x} \in \ts{X}'$ is assigned a rank $\alpha$ based on the  non\hyp{}dominated sorting.
A secondary rank $\beta$ is assigned to each solution based on the value of a hyper-volume metric~\cite{glasmachers2010natural,zitzler1998multiobjective} in a descending order.
To calculate the hyper\hyp{}volume metric, a dominated reference point $\tv{\zeta}^{(0)} \in \mathbb{R}^M$ is selected in the objective space, such that all considered solutions $\tv{x} \in \ts{X}'$ dominate this point $\tv{\zeta}(\tv{x}) \prec \tv{\zeta}^{(0)}$.
The hyper-volume metric~\cite{zitzler1998multiobjective} is used to calculate the hyper-volume between each solution and the dominated reference point, e.g. by using the proposed implementation of Ref.~\cite{fonseca2006improved}.
The hyper-volume metric is calculated on normalized loss values, which are calculated by subtracting the mean and then dividing with the standard deviation over all solutions.
The covariance related parameters $\matr{A},\sigma$ are updated with a gradient update.
A modified version of MO-NES, where solutions are rounded and negative values are clipped to 0 prior to evaluation is used as a baseline in the current article.
The initial value of each solution is sampled as $x_i = \text{ReLU}(x), x \sim \mathcal{N}(0, 0.2)$.
Parameter $\sigma = 1/3$ and elements of $A$ were initialized uniformly in $[0, 0.001]$.
Following notation from Ref.~\cite{glasmachers2010natural}, the learning rates for each parameter are $\eta_\sigma^+ = 0.01$,  $\eta_\sigma^- = 0.01/5$ and $\eta_A = 0.01/4$.
MO-NES trains up to 40 generations.

\subsection{Gradient Guided Genetic Algorithm}\label{section:methods:g3}

\begin{figure}
    \centering
    \subfloat[GA\label{fig:g3:ga}]{\includegraphics[width=0.40\textwidth,valign=t]{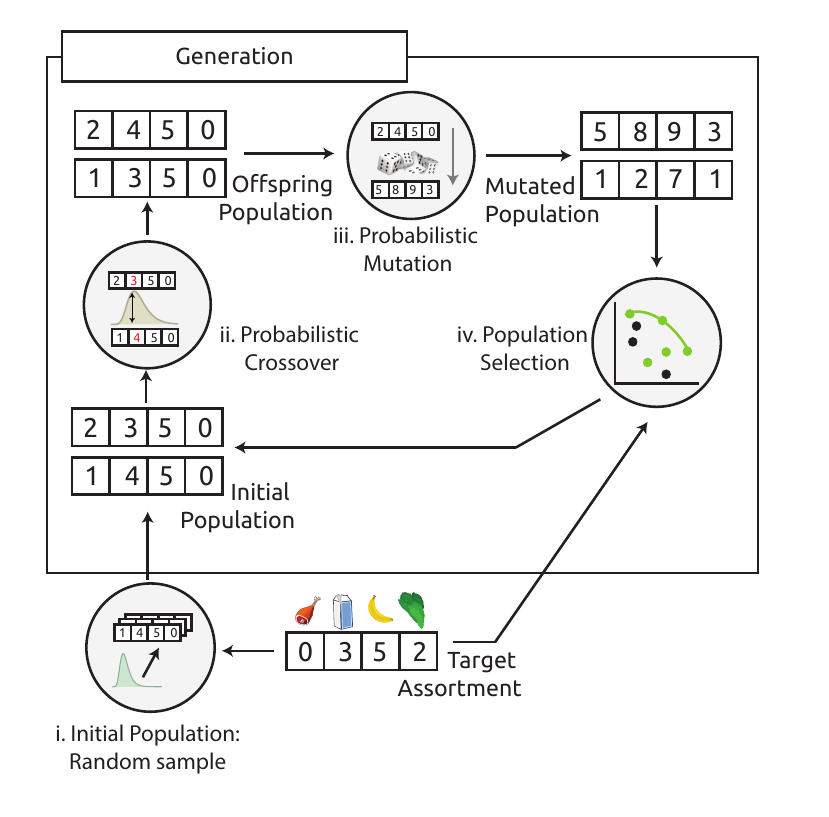}}
    \subfloat[G3A\label{fig:g3:g3a}]{\includegraphics[width=0.58\textwidth,valign=t]{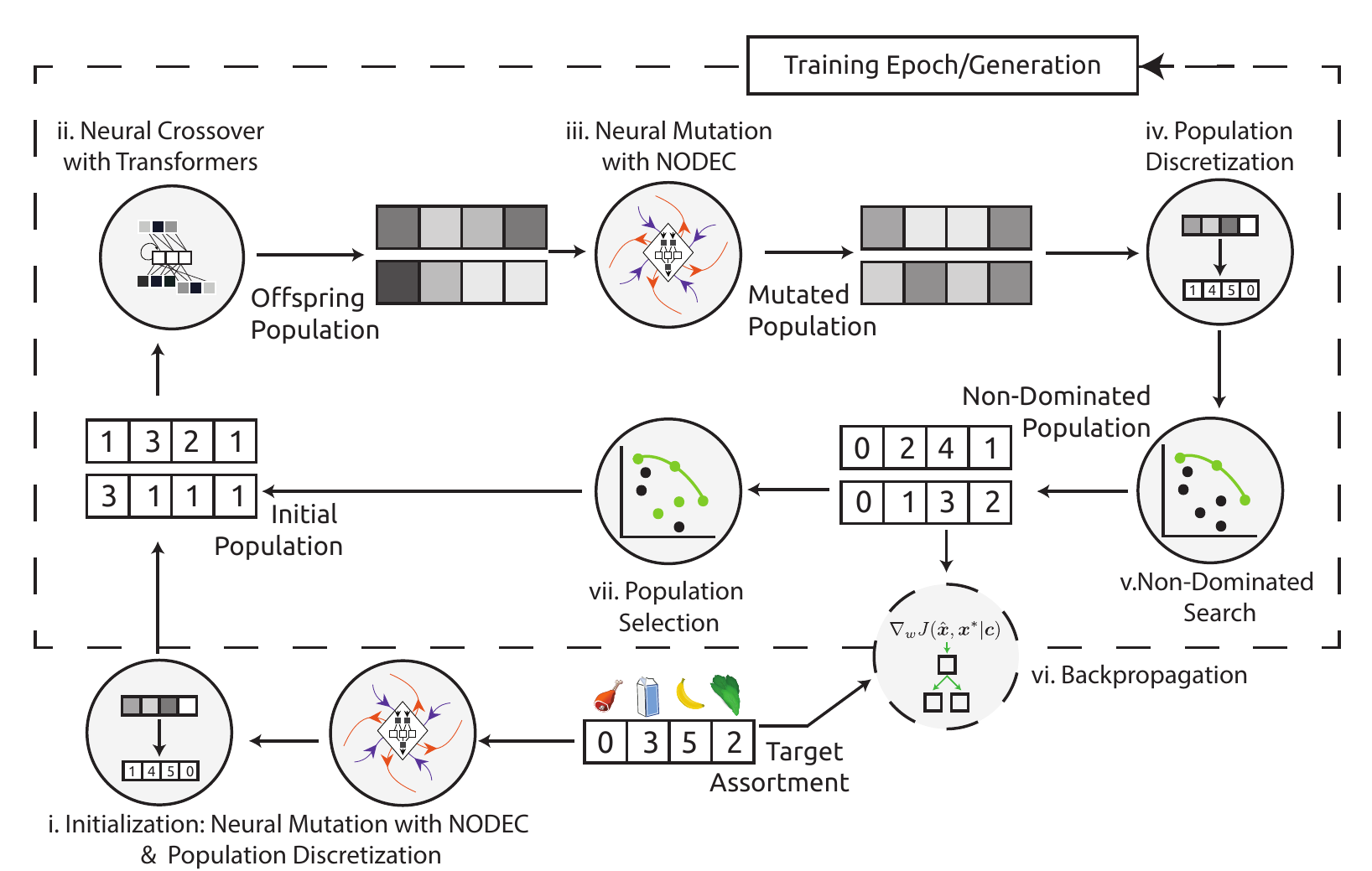}}

    \caption[A high level illustration of the G3A model vs a classic GA.]{A high level illustration of the G3A and the classic GA. In the classic GA (a), the components of initial population sampling (a.i), crossover (a.ii), and mutation (a.iii) are often performed in a probabilistic manner.
    The population selection (a.iv) applies various mechanisms, such as a non\hyp{}dominated sorting, over the current population to select the solutions that will continue in the next generation.
    Often a subset of the non\hyp{}dominated solutions is preserved to the next generation.
    In G3A (b.) the initial population sample (b.i), the crossover (b.ii), and mutation operators (b.iii) are performed with a neural network. 
    A discretization technique that allows back-propagation is then applied (component b.iv), and then a gradient descent update is performed based on the loss values of all non\hyp{}dominated solution (components b.v-b.vi). Population selection (b.vii) happens after the network update, thus gradient guided genetics preserves information from all non\hyp{}dominated solutions in its weight.
    }
    \label{fig:g3}
\end{figure}

Probabilistic algorithms may suffer from slow convergence~\cite{ishibuchi2009evolutionary}, especially on high dimensional problems.
Dependence on randomness and selection of random seeds may also be considered as a challenge~\cite{fuertes2019chaotic,lu2014effects}.
Recently, deterministic chaos genetic algorithms have been proposed to calculate solutions in a deterministic and seemingly in a more efficient manner~\cite{yan2003chaos,yong2002study}.
Furthermore, chaotic maps seem very promising for sparse and highly dimensional problems as they can control entropy~\cite{fuertes2019chaotic} and the performance of the optimization procedures.
For example, genetic algorithms may show improved performance if a logistic map~\cite{fuertes2019chaotic} is used to sample initial solutions around the intended basket.
Nevertheless, chaos genetic algorithms do not use explicit feedback from the loss function, such as MO-NES~\cite{glasmachers2010natural}, and often the selection of adequate chaotic maps requires extensive hyper-parameter optimization~\cite{fuertes2019chaotic,lu2014effects}.
This article investigates another potential design, where neural networks are used to perform \emph{mutation and crossover} operators and/or \emph{initialize the population} instead of chaotic maps.
Neural networks show promising capabilities to learn chaotic maps and strange attractors~\cite{li2021neural}, and back-propagation can be used to learn the parameters of the neural networks and control the chaotic behavior to improve solutions across generations.
In this article a novel gradient guided genetic algorithm is proposed by combining design concepts from chaotic genetic algorithms and neural networks.
G3A may evolve a population of solutions conditional to input data (such as the coefficient matrix) by performing gradient guided genetic operations.
An overview of G3A is provided based on \cref{fig:g3:g3a}.

\subsubsection{Population Initialization}
An initial population matrix $\matr{X}_0$ is calculated by applying the untrained neural mutation from $t=0$ to $t = T$.
$B$ solutions are selected during initialization, by sampling the mutation trajectory every $\Delta t = T/B$.
During each generation, a population matrix $\matr{X}_\tau \in \mathbb{N}_0^{B\times N}$ is created, where each row represents a recommended solution.

\subsubsection{Neural Crossover}
A neural crossover operator is then applied on the population matrix and generates an offspring solution for each solution in the initial population.
The main neural network component is a transformer network $f_{\text{transformer}}: \mathbb{N}_0^{B \times N} \to \mathbb{R}^{B \times B \times N}$ with Gaussian Error Linear Unit~\cite{hendrycks2016gaussian} (GeLU) activation functions as hidden layers~\cite{vaswani2017attention}.  
Each parent solution $\tv{x}$ is compared with the rest of the population matrix  $\matr{X}_\tau$.
For each element $x_i$ of the parent solution, the transformer generates an attention vector $\tv{g} \in \mathbb{R}^B$ over all solutions in the population.
The element $\hat{x}_{i,b}$ is selected from the $b\text{-th}$ parent in the population that received the maximum attention value from the transformer $b = \argmax_b g_b$.

The multi-head attention transformer networks used in this work contain 1 encoder and 1 decoder layer with GeLU activation functions and 11 heads.
Replacing the crossover network with probabilistic operators or simpler neural network architectures has not yielded better results so far, but is still a subject of study and future work.
Both of the transformer encoder and decoder layers contain a hidden layer with 2048 hidden neurons and, layer norm layers in output and input and also dropout operations on neuron outputs\footnote{according to the default implementation found in \url{https://pytorch.org/docs/stable/generated/torch.nn.Transformer.html} (accessed October 2021)}.
A sigmoid activation is then applied on the attention values and each selected parent element is used to calculate an ``offspring'' solution element $x_i'$ in the following manner:
\begin{equation}\label{eq:neural:crossover}
    x_i' =  \text{sigmoid}(g_k) x_i + (1-\text{sigmoid}(g_b)) \hat{x}_{i,b} .
\end{equation}

\subsubsection{Neural Mutation}
Next, a mutation operator neural network $u(\boldsymbol{x}(t)): \mathbb{R}^{B\times N}\to \mathbb{R}^{B\times N}$ evolves a solution $\boldsymbol{x}(t)$ in continuous time $t$ by applying the following neural ODE control
\begin{equation}\label{eq:neural:mutation}
    \dot{\boldsymbol{x}}(t) = \boldsymbol{u}(\boldsymbol{x}(t)).
\end{equation}
A neural ODE solve~\cite{chen2018neural} scheme is used to calculate the continuous time evolution between subsequent genetic generations, e.g. $\boldsymbol{x}(0) \to \tv{x}(T)$.
The underlying neural network has sinusoidal activation functions in the hidden layer, inspired by the sinusoidal iterator used in Ref.~\cite{lu2014effects}.
The evolution period is $[0,1]$ and a single hidden layer with $256$ neurons.
To select the solutions that are preserved to the next generation, a finite number of mutated solution is sampled uniformly across time for each solution $\tv{x}(0)$ of the current generation $\tau$ at predetermined time-steps within the ODE solver. 
The output activation of the neural network is a Rectifier Linear Unit (ReLU) activation~\cite{vaswani2017attention}, which removes negative product quantities from each solutions.

Neural network weights and activation functions generate real\hyp{}valued solutions. The proposed problems require discrete product quantities in the solution. 
Therefore, a discretization operation that allows gradient propagation is applied on each solution.
G3A can be viewed as an extension of neural ODEs control ~\cite{asikis2020nnc,bottcher2022ai} with discrete events~\cite{chen2020learning} to MIP problems.

\subsubsection{Fractional Decoupling}\label{supp:fd}
Neural networks are known to operate in real value settings, as back-propagation requires the output of neural networks to be continuous and differentiable in regards to objective, 
so that the chain rule can be efficiently applied.
Continuous outputs are not compatible with end-to-end learning mixed integer programming problem (MIP) solutions.
Rounding neural network outputs creates a challenge when back-propagating error for the calculation of the gradient, as rounding functions are not differentiable in its domain, in particular at the integer values. %

A potential approach is to train the neural networks in a real valued manner and then apply rounding when evaluating the solutions, e.g. applying a linear programming relaxation scheme~\cite{klotz2013practical,patro2020towards}.
When considering shopping baskets over a wide variety of products, such relaxations may become problematic.
Neural networks with many outputs may assign a small positive quantity over hundreds of products to a single basket to optimize taste and environmental losses.
In such case many product quantities are rounded to $0$, yielding empty baskets or very sparse baskets as solutions.
Another approach proposed in literature is to use the Gumbel soft-max operator~\cite{jang2016categorical}, which allows for gradient propagation via the aforementioned straight\hyp{}through estimators~\cite{yin2019understanding}.
Since the decision problem in question requires no upper bounds on purchased product quantities, using the Gumbel soft-max operator may yield high dimensional outputs that may require more time to train for large scale problems.

An alternative approach, termed fractional decoupling, is proposed to efficiently calculate a gradient update and perform gradient descent via a straight\hyp{}through estimator~\cite{bengio2013estimating,yin2019understanding}.
To perform fractional decoupling, one subtracts the fractional part $h_i$ of a real valued output $y_i$, while treating the fractional part as constant, i.e. this allows no gradient propagation through the fractional part in the computational graph.
This operation can be considered as a rounding straight-though estimator.

\subsubsection{Selection}
A non\hyp{}dominated sorting is performed across all discretized solutions to determine the best solutions from each trajectory.
The mean objective value per feature  
\begin{equation}
    \overline{\zeta}_j = \dfrac{\sum_{\tv{x}(t) \in F_1} \zeta_j(\tv{x}(t))}{|F_1|}  
\end{equation}
is calculated over all non\hyp{}dominated solutions, i.e. all samples $\tv{x} \in F_1$, and then each element $\overline{\zeta}_j$ is used to calculate gradients and perform the parameter update.
Mean objective values $\overline{\zeta}_j$ can be scaled before gradient calculation to match consumer preferences and guide the algorithm towards non\hyp{}dominated solutions that are better performing in specific objective values.
To select the $B$ solutions that are used as input population for the next generation, the hyper-volume and non\hyp{}dominated ranks are used~\cite{glasmachers2010natural} as described in the MO-NES baseline in \cref{chapter:appendix:g3:MONES}.

\subsubsection{Back-Propagation Through Evolution}
The back-propagation through evolution starts by calculating the individual objective function values for each solution selected by the selection operator.
For each objective, the mean value over all selected individuals is calculated.
Experimental results indicated that using a different optimizer for each Neural Operator yields higher performance. 
The RMSProp optimizer is used with learning rate $\eta = 0.0001$ for the neural mutation operator and an RMSProp optimizer with learning rate $\eta = 0.0001$ for the Neural Crossover Operator.
The gradient is calculated iteratively per objective, and for health objectives the loss is scaled $7$ times.
Such scaling resembles a scalarization method~\cite{ZHENG2022141} optimization, although a weighted sum may not used for gradient calculation.
Not scaling the loss yielded recommendations that did not optimize health objectives well, as environmental and cost objectives were often positive-correlated and dominated the gradient upgrades.
In general, each objective loss can also be scaled to match explicit consumer preferences.
Consumers may rank or score most important objectives, and such scores can be used as scaling factors for the objectives~\cite{asikis2020value}.

\section{Experimental Evaluation}\label{sec:g3a:experimental:evaluation}

Two multi-objective optimization algorithms are compared with G3A, namely MO-NES, and reference point NSGA-II~\cite{deb2006reference} (RNSGA-II).
All baselines are evaluated in  weekly basket purchases that happen over the course of 85 weeks for 500 households, and in total 28400 intended baskets are considered.
In particular, the households are chosen based on their total green house gasses (GHG) emissions, i.e. the top 500 emission producers are selected.
G3A is parameterized to generate $B=8$ recommendations per intended basket, whereas RNSGA-II and MO-NES generate $B=10$ recommendations per intended basket.
The population sizes where chosen after evaluating different values.
The sizes that generated well-performing solutions efficiently were preferred.
For each recommendation, a ratio towards the cost, environmental impact, or nutritional quantities of the intended basket are considered.
Some of the ratio functions coincide with the proposed objective functions, but this is not the case for nutritional losses, as the normalized MSE showed better convergence, but required scaling.
Other GA baselines were also considered~\cite{nsga3,deb2002fast,rnsga3}, but did not produce competitive results and thus are skipped for the sake of brevity.
Further technical details on hyper-parameters and experimental setup are also found in the code accompanying the article.

It is important to note that all three baselines were tested on a subset of potential hyper-parameters. Hyper-parameter optimization was performed for several days to the extend that each method was able to solve the problem effectively. From observed models, the best performing parameterization per method was selected. In future work, G3A will be compared against other optimization methods on more established problems to determine performance in terms of optimality. Such a study was out of the scope of this article.

\subsection{Recommendation Comparison}

First, the ability of baselines to produce non\hyp{}dominated solutions for the problem is evaluated.
\cref{tab:g3a:dominance} contains a comparison where all recommendations for an intended basket $\tv{x}^*$ from all methods are compared against each other and only the non\hyp{}dominated solutions are kept across all methods.
The ratio of total non\hyp{}dominated solutions divided by total recommendations per method is calculated.
All three baselines produce diverse non\hyp{}dominated solutions, as they all achieve high mean ratio of non\hyp{}dominated to total recommended baskets per intended basket.
This indicates that the problem can be tackled effectively by all methods.

\begin{table}[!htb]
    \centering
    \small
    \begin{tabular}{lrrrr}
    \toprule
    Model &      Mean &  Mean CI &  Median &   Median CI \\
    \midrule
    G3A  &  0.980 &  (0.979, 0.981) &     1.0 &  (1.0, 1.0) \\
    MO-NES  &  0.948 &  (0.946, 0.949) &     1.0 &  (1.0, 1.0) \\
    RNSGA-II &  0.986 &  (0.985, 0.986) &     1.0 &  (1.0, 1.0) \\
    \bottomrule
    \end{tabular}
    \caption[Non\hyp{}dominated solution analysis across all baselines.]{
    Mean and median values of non\hyp{}dominated percentage of solutions when recommendations from all methods are combined together.
    Reverse bootstrap confidence intervals with significance level $\alpha=0.05$ are also provided.
    All three baselines find a high percentage of non\hyp{}dominated solutions, even when compared to each other.
    }
    \label{tab:g3a:dominance}
\end{table}

\begin{figure}[!htb]
    \centering
    \subfloat[Cost and Environmental Impact Ratios (lower values - points closer to center are preferred).]{\includegraphics[width=0.45\textwidth]{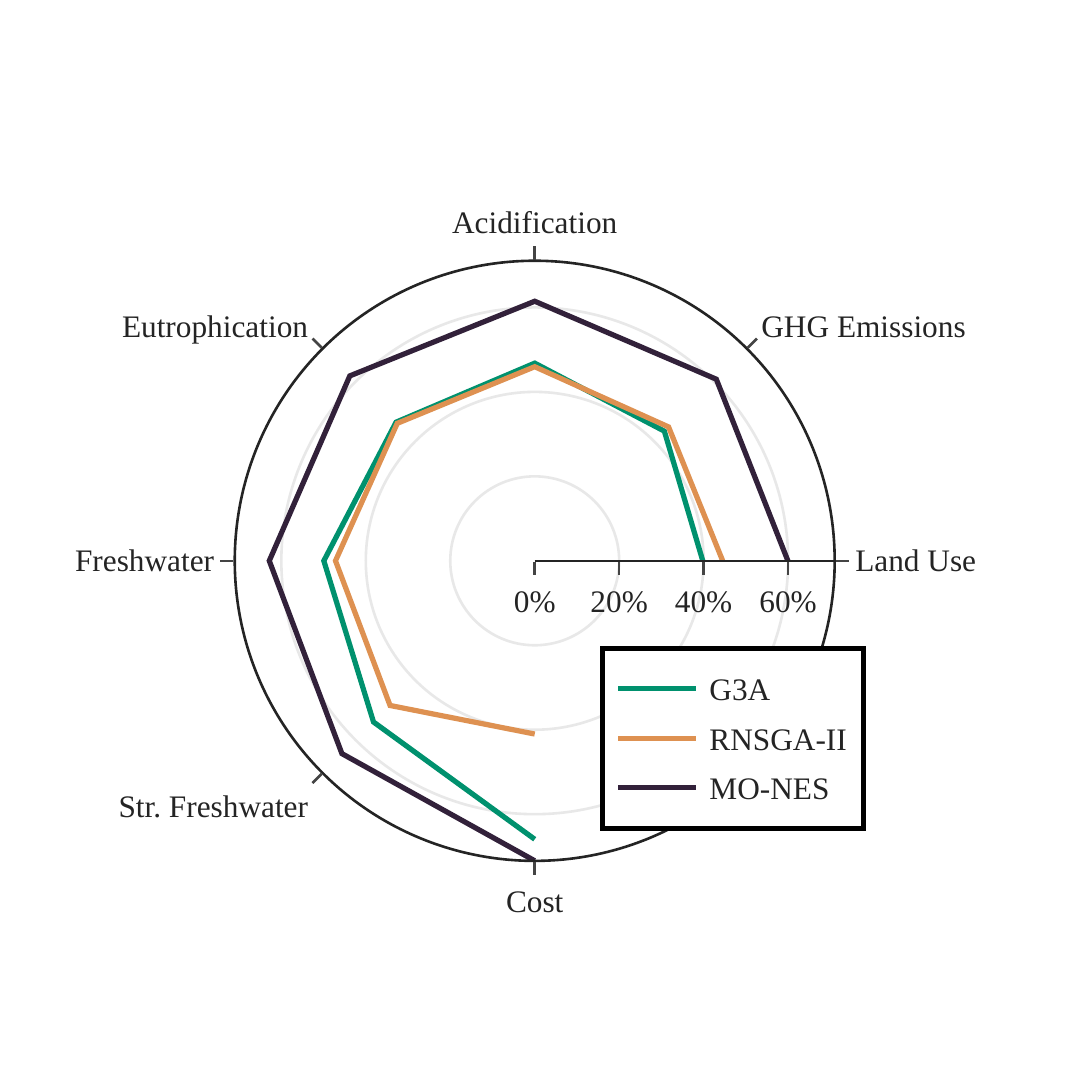}}
    \hfill
    \subfloat[Nutritional ratios and cosine similarity (higher values - longer bars are preferred).]{\includegraphics[width=0.48\textwidth]{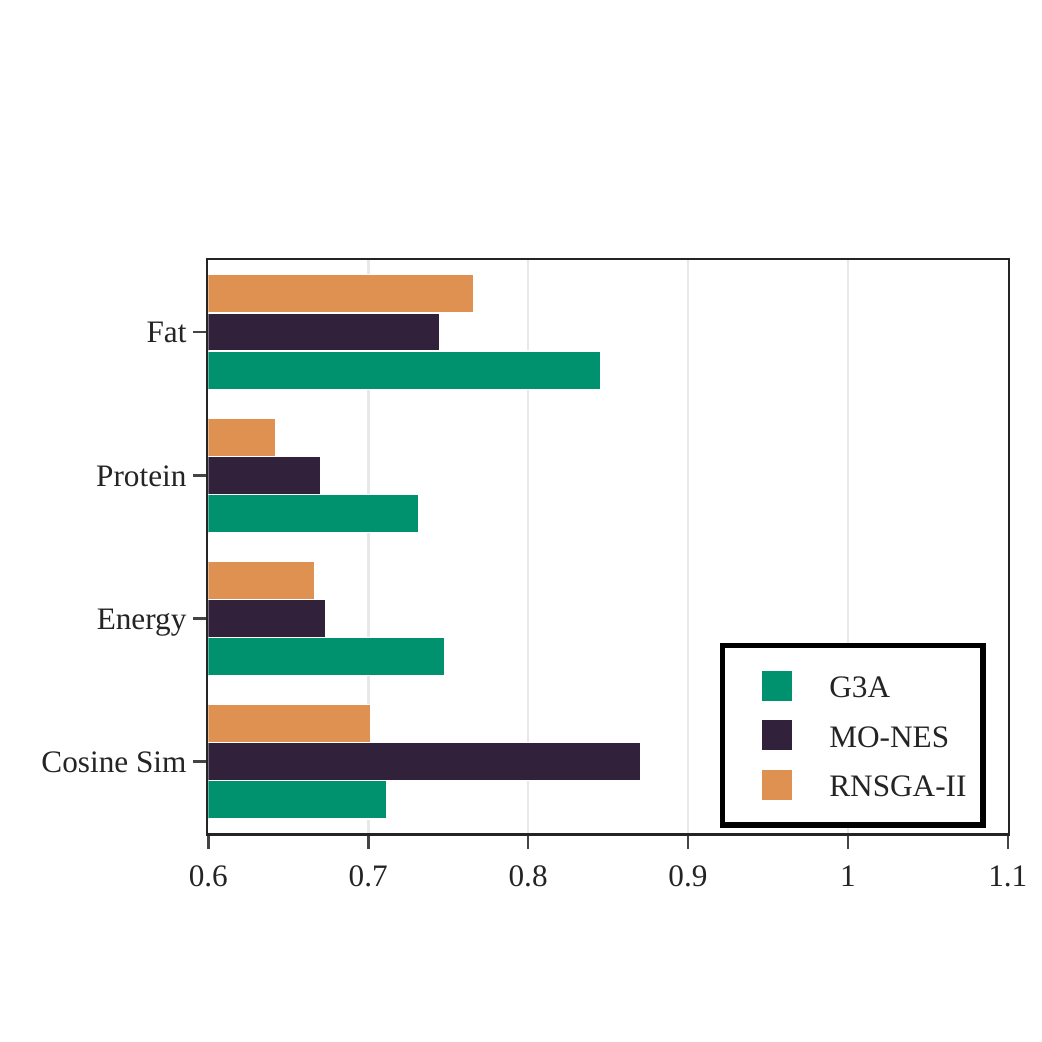}}
    \caption[Mean recommendation relative performance comparison.]{
    A comparison of cosine similarity and the total emission, nutritional, and cost of a recommendation, as a ratio to the corresponding intended basket.
    For each baseline the mean ratio value over all recommendations that achieve cosine similarity higher than $0.5$ and have all environmental ratios costs below $1.0$ are considered.
    }
    \label{fig:g3a:means}
\end{figure}

Several recommended baskets per model may have non-preferred objective values.
For example, a solution may achieve the optimal value in terms of a nutritional loss and then be selected as a non\hyp{}dominated solution, although it produces $200\%$ more emissions.
Such solutions are discarded for the comparisons in the next sections, i.e. not recommended.
A filtering is applied by discarding any solution that has a cost or any environmental impact quantity or cost ratio $\rho({\tv{x},\tv{x}^*}) \geq 1.0$.
Furthermore, very dissimilar baskets are also discarded, i.e. when $\text{cosine\_sim}(\tv{x}, \tv{x}^*) \leq 0.5$.
For each recommendation the cosine similarity and environmental impact, nutritional and cost ratios towards the corresponding intended basket are calculated.
The mean value of the ratio calculations over all recommendations per model are reported in \cref{fig:g3a:means}.
\cref{fig:g3a:means} indicates that RNSGA-II outperforms other baselines in terms of cost, while G3A shows higher performance in terms of nutritional values.
MO-NES shows higher performance in terms of cosine similarity.

As indicated by the dominance analysis results, all models can provide highly dominant solutions, that potentially specialize better in subsets of objectives.
Depending on the design goals of the system or the priority of the individual, a different algorithm might be more preferred.
Furthermore, all three models can be further altered to include consumer input in which objectives need to be prioritized. 
For G3A and MO-NES this can be implemented by adding weights and scaling the objective function values before the gradient update.
For RNSGA-II this can be achieved by creating reference points that correspond to the consumer priorities.

\subsection{Calculation Execution Time and Emissions}

A sample of $100$ intended baskets over a single week is used to determine execution time and calculation GHG emission for each model (see \cref{tab:g3a:timesemisssion}).
Although G3A requires higher computation time and generates more emissions per calculation of recommendations, all models produce emissions and wall clock times are not significant.
Accepting a single recommendation of any model can justify the emissions of thousands or even millions of calculations of other recommendations.
Furthermore, G3A code is still at an experimental stage, and better code optimization can be achieved to further reduce calculation times and emissions. 
One possible way to achieve such improvements on G3A would be to reduce the number of artificial neurons in the Neural Crossover transformer network.

\begin{table}[!ht]
    \footnotesize
    \begin{tabularx}{\hsize}{L{0.13\hsize}R{0.15\hsize}R{0.2\hsize}R{0.35\hsize}}
    \toprule
    Model &  Elapsed Wall Clock Time &  GHG Emissions &  Mean, Min GHG Emission Improvement  \\
    {} & seconds  & kg CO\textsubscript{2} eq. &  kg CO\textsubscript{2} eq. \\
    \midrule
    G3A  (GPU)      &      1.89 $\pm$ 1.22   &    2.07 $\pm$  1.44)e-08  &  31.49, 0.46     \\
    MO-NES (CPU)    &      0.20 $\pm$ 0.01   &    (2.16$\pm$ 0.14)e-09   &  21.03, 0.41   \\
    RNSGA-II (CPU)  &      0.46  $\pm$  0.06 &    (6.95$\pm$2.41)e-10    &  34.04, 0.45    \\
    \bottomrule
    \end{tabularx}
    \caption[Baseline execution times and emissions.]{Execution time and GHG emissions (mean $\pm$ standard deviation) measured with python and the codecarbon library~\cite{codecarbon} over a sample of 100 intended baskets from different households. 
    The mean and minimum GHG emission improvement for accepting a single recommendation is also reported to outline the potential cost-benefit of accepting versus calculating recommendations.}
    \label{tab:g3a:timesemisssion}
\end{table}

\subsection{Real-World Impact}

\begin{figure}
    \centering
    \subfloat[Mean nutritional value and cost per basket per trajectory.\label{fig:g3a:means:nutr}]{\includegraphics[width=0.48\textwidth]{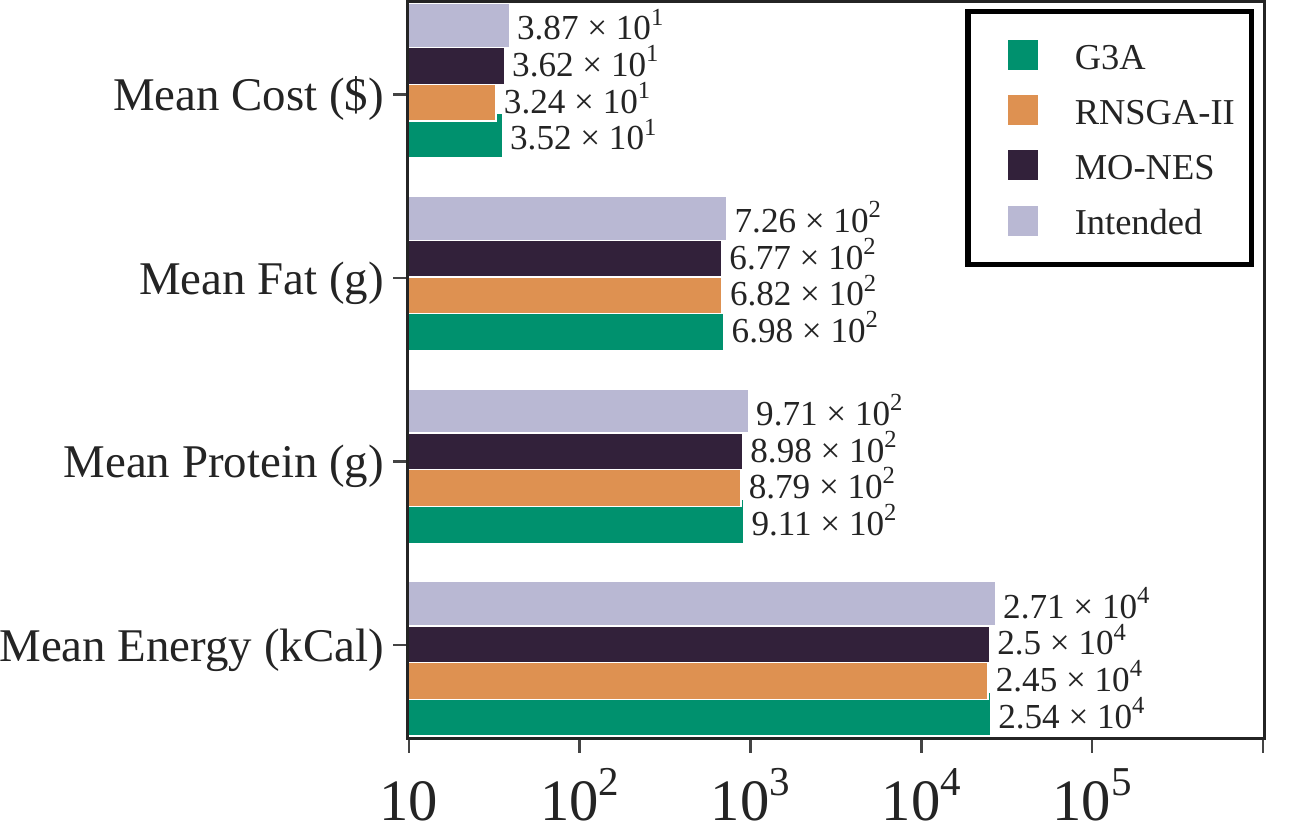}}
    \hfill
    \subfloat[Mean ratios of added and removed units per basket per trajectory.]{\includegraphics[width=0.48\textwidth]{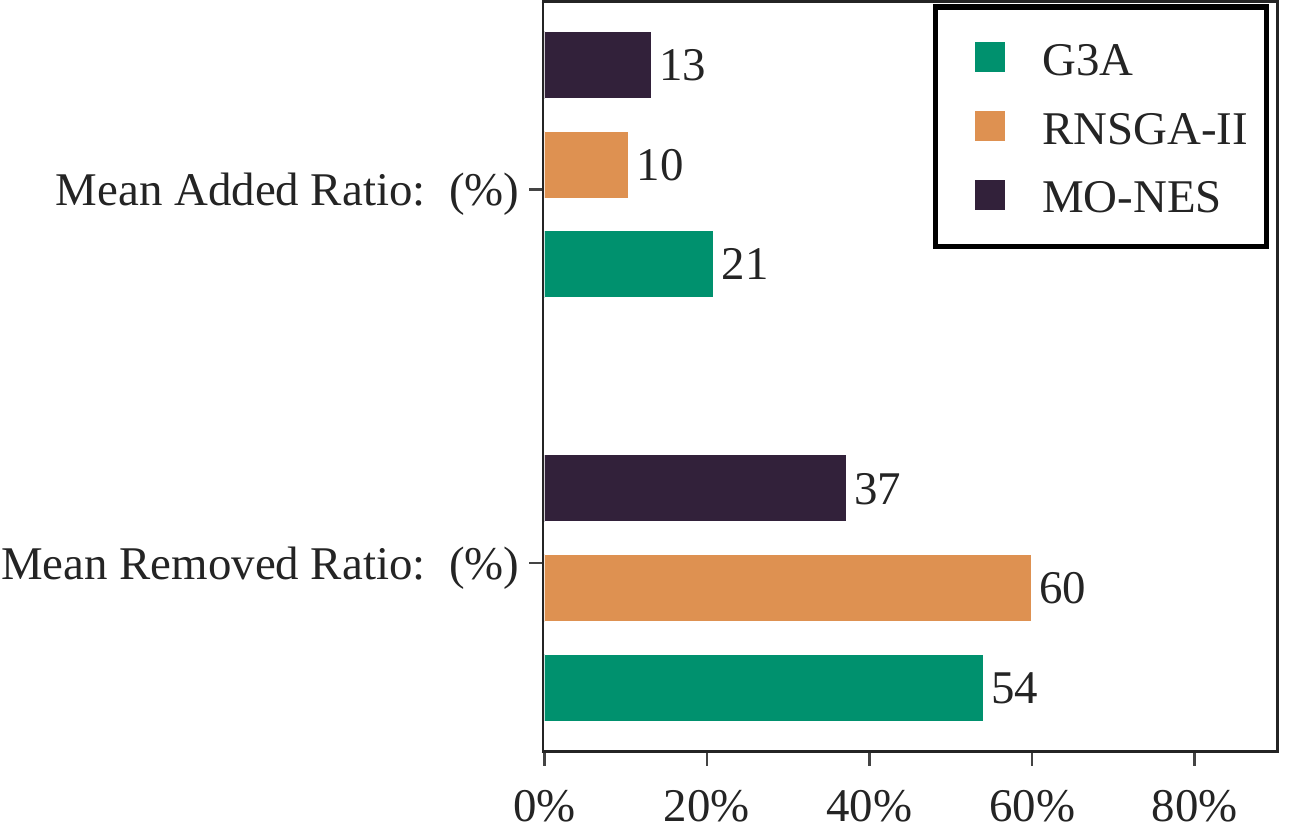}}\\
    \subfloat[Mean reduction of total environmental impact of accepting recommendations versus purchasing only intended baskets per sampled trajectory. Longer bars perform better.]{\includegraphics[width=0.98\textwidth]{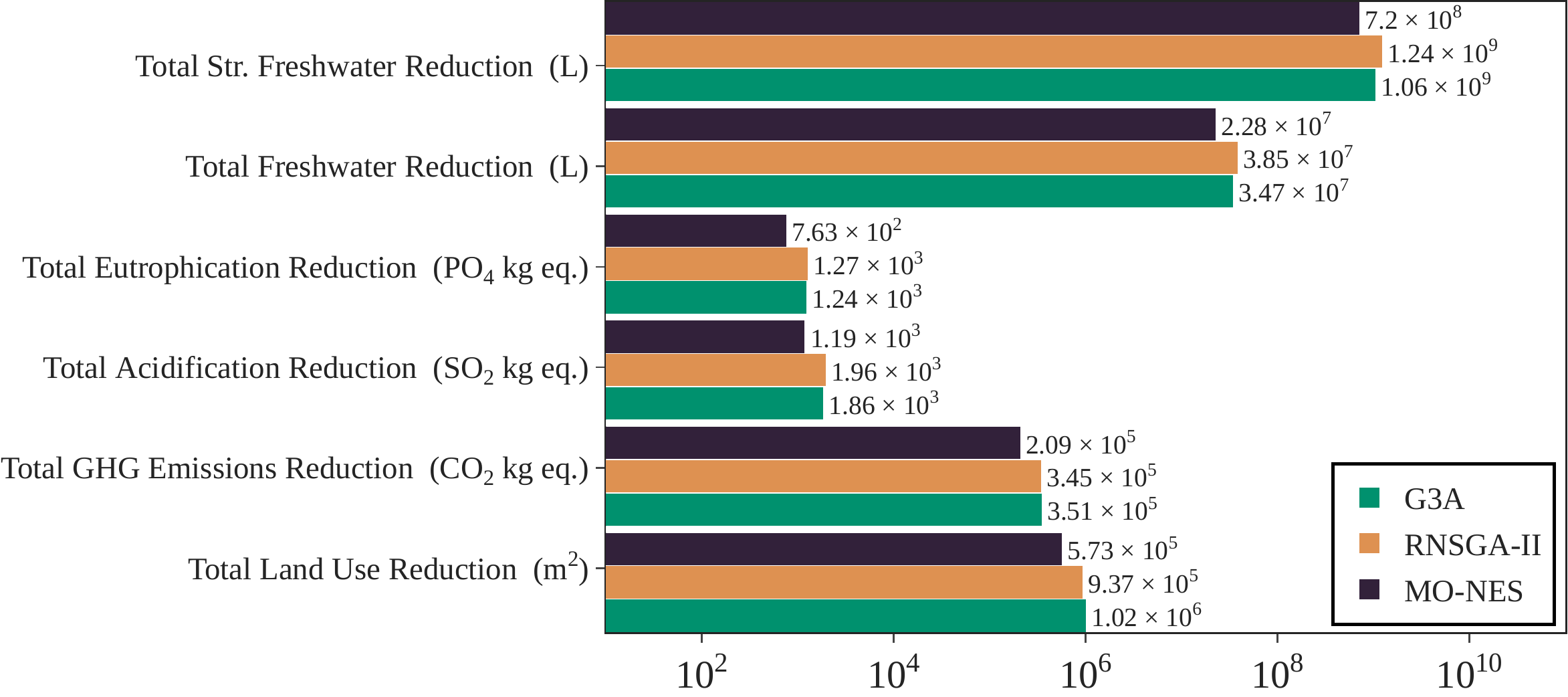}}
    \caption[Mean recommendation impact comparison.]{Comparison of the impact per model over $5000$ trajectories, where $25\%$ of the intended basket purchases is randomly replaced with a recommendation.
    Mean nutritional quantities per basket and trajectory are reported (see a.).
    Next (see b.), the mean value of added and removed units per basket are provided over all recommendations and trajectories, where intended baskets are omitted for the calculation.
    The total environmental impact and cost reduction are calculated per sample and then subtracted from the total quantities of the original trajectory (only intended baskets are purchased).
    Although confidence intervals are calculated, they are omitted as they are mostly too narrow and, thus, not visible.
    }
    \label{fig:g3a:means:samples}
\end{figure}

To extend the comparison of G3A and estimate the impact on total reduction values, a counterfactual scenario is evaluated.
For each model, $5000$ counterfactual trajectories are sampled, each trajectory being $86$ weeks long.
For each trajectory, it is assumed that $25\%$ of all intended baskets are replaced with a recommendation.
The recommendation which replaces the intended basket is chosen randomly\footnote{In the current setting, a decision model for sampling, such as the one in \cite{kwak2015analysis} cannot be used, because the transactions of the current dataset may be effected by marketing campaigns and other covariates.
Furthermore, it is not apparent of whether consumers were aware of sustainability issues when performing a purchase, thus the modeling of environmental impact decision factors may be invalid.
Thus, designing a valid decision model to estimate the effect of a recommender system in this case is out of scope of this article and could potentially be considered as future work.
}.
\cref{fig:g3a:means:samples} illustrates the ability of all algorithms to achieve a considerable reduction of environmental impact compared to the intended basket.
For example, deciding to replace $25\%$ of intended baskets with a G3A recommendation leads to a reduction of approximately 35 metric kilo-tons of CO\textsubscript{2} eq. or approximately 1 billion litres of stressed freshwater for G3A.
The current results indicate that G3A achieves similar performance to RNSGA-II, but by removing less and adding more products.
MO-NES instead produces recommendations that have the least impact on the consumer basket.

\section{Conclusion}\label{sec:g3a:conclusion}

This article showcases a multi-objective approach to sustainable recommendations, where value\hyp{}sensitive design is also taken into account.
The problem of finding sustainable personalized baskets is evaluated with objectives and constraints derived from real\hyp{}world data.
Existing baselines are compared with a novel gradient guided genetic algorithm and results showcase that all considered models produce good solutions to the problem.
Even when individuals would adopt a fraction of the sustainable recommendations, a considerable environmental impact can be observed.

From a technical perspective this paper introduces a novel multi-objective optimization algorithm (G3A) that achieves comparable performance with state-of-the-art baselines on the new task.
To make this happen, existing techniques from evolutionary methods are combined with state-of-the art neural network architectures, such as Neural ODE Controllers, attention and a straight-though estimator for discretization, termed fractional decoupling.
These outcomes may be further tested in future work as multi-objective optimization baselines in other settings as well.

\section{Acknowledgements}
This work/paper has been partially supported by the 'Co-Evolving City Life - CoCi' project through the European Research Council (ERC) under the European Union’s Horizon 2020 research and innovation programme (grant agreement No. 833168).
The author would like to acknowledge the NCCR Automation for the support.
Finally, the author would also like to thank Dr. Nino Antulov-Fantulin, Dr. Lucas Boettcher, Prof. Dirk Helbing, Prof. Petros Koumoutsakos and Prof. Avished Anand for their inputs and comments on the manuscript.

\clearpage
\printbibliography %

\end{document}